\documentclass[conference]{IEEEtran}
\IEEEoverridecommandlockouts
\usepackage{cite}
\usepackage{amsmath,amssymb,amsfonts}
\usepackage{algorithmic}
\usepackage{graphicx}
\usepackage{textcomp}
\usepackage{xcolor}

\usepackage{booktabs}
\usepackage{algorithm2e}
\usepackage{placeins}
\usepackage{float}
\usepackage{soul}
\usepackage{xcolor}

\def\BibTeX{{\rm B\kern-.05em{\sc i\kern-.025em b}\kern-.08em
    T\kern-.1667em\lower.7ex\hbox{E}\kern-.125emX}}

\begin{document}

\title{Enhancing Classification with Semi-Supervised Deep Learning
       Using Distance-Based Sample Weights\\[0.3em]
       \thanks{This paper has been accepted for publication and oral presentation at the 2025 10th IEEE International Conference on Machine Learning Technologies (ICMLT 2025). The final authenticated version will be available in IEEE Xplore following the conference.}}

\author{
\IEEEauthorblockN{
\begin{minipage}[t]{0.32\linewidth}\centering
Aydin Abedinia\\
\textit{Dept. of Computer Science}\\
\textit{Islamic Azad University South-Tehran Branch}\\
Tehran, Iran\\
abedinia.aydin@icloud.com
\end{minipage}
\begin{minipage}[t]{0.32\linewidth}\centering
Shima Tabakhi\\
\textit{School of Electrical, Computer and Biomedical Engineering}\\
\textit{Southern Illinois University}\\
Carbondale, United States\\
Shima.tabakhi@siu.edu
\end{minipage}
\begin{minipage}[t]{0.32\linewidth}\centering
Vahid Seydi\\
\textit{School of Ocean Sciences}\\
\textit{Bangor University}\\
Menai Bridge, UK\\
v.seydi@bangor.ac.uk
\end{minipage}
}
}

\maketitle

\begin{abstract}
Recent advancements in semi-supervised deep learning have introduced effective strategies for leveraging both labeled and unlabeled data to improve classification performance. This work proposes a semi-supervised framework that utilizes a distance-based weighting mechanism to prioritize critical training samples based on their proximity to test data. By focusing on the most informative examples, the method enhances model generalization and robustness, particularly in challenging scenarios with noisy or imbalanced datasets. Building on techniques such as uncertainty consistency and graph-based representations, the approach addresses key challenges of limited labeled data while maintaining scalability. Experiments on twelve benchmark datasets demonstrate significant improvements across key metrics, including accuracy, precision, and recall, consistently outperforming existing methods. This framework provides a robust and practical solution for semi-supervised learning, with potential applications in domains such as healthcare and security where data limitations pose significant challenges.
\end{abstract}


\section{Introduction}
Neural networks (NN) and deep learning (DL) have transformed classification tasks across domains such as image classification \cite{chum2017cnn}\cite{affonso2017deep}\cite{obaid2020deep}\cite{wang2021comparative}, natural language processing \cite{torfi2020natural}\cite{young2018recent}, medical diagnosis, face recognition \cite{guo2019survey}\cite{fuad2021recent}, and object detection \cite{hu2015face}\cite{alzubaidi2021review}. However, despite their success, modern DL models critically depend on large-scale, well-annotated datasets \cite{krizhevsky2012imagenet}\cite{chen2021attention}, which are both expensive and labor-intensive to produce. This dependency often leads to issues such as overfitting and limits the deployment of DL techniques in scenarios where labeled data is scarce.

Semi-supervised learning (SSL) has emerged as a promising alternative by effectively leveraging both labeled and unlabeled data to improve model generalization. Existing SSL approaches---such as self-training \cite{he2011self}\cite{khezri2020stds}, graph-based methods \cite{sawant2020review}\cite{subramanya2022graph}, and pseudo-labeling techniques \cite{oh2022daso}---demonstrate that incorporating unlabeled data can substantially alleviate the need for extensive labeling. Nonetheless, many of these methods are susceptible to error propagation, particularly when confronted with class imbalance and noisy labels.

In this paper, we propose a state-of-the-art classification framework that employs a distance-based weighting mechanism to construct a training set that is both efficient and accurate. By assigning continuous weights based on the similarity between training samples and the target distribution, our method minimizes the dependency on large labeled datasets and mitigates the adverse effects of noise and imbalance \cite{abedinia2024building}. Our contributions are threefold:
\begin{itemize}
    \item We introduce an adaptive distance-based weighting strategy that prioritizes informative samples during training.
    \item We integrate these weights into the deep neural network’s loss function to guide gradient updates more effectively.
    \item We validate our approach on benchmark datasets, demonstrating improvements in robustness, generalization, and classification accuracy.
\end{itemize}

\section{Related Work}

Recent advances in semi-supervised learning have significantly enhanced our ability to leverage unlabeled data, addressing challenges such as data scarcity and class imbalance. Several key directions in the literature are particularly noteworthy.

\paragraph{Self-Training and Pseudo-Labeling}  
Self-training frameworks that iteratively generate pseudo-labels for unlabeled data have been widely explored \cite{he2011self}\cite{khezri2020stds}. Although these approaches have improved performance in many scenarios, they remain vulnerable to error propagation—especially when initial pseudo-labels are noisy or imbalanced. Youngtaek Oh et al. \cite{oh2022daso} propose a distribution-aware pseudo-labeling strategy to mitigate these risks, yet the inherent challenges in balancing errors persist.

\paragraph{Graph-Based and Generative Models}  
Graph-based SSL methods have attracted considerable attention for their ability to capture the intrinsic structure of data \cite{sawant2020review}\cite{subramanya2022graph}. Similarly, generative models have been employed to model the distribution of unlabeled data effectively \cite{kingma2014semi}. Despite their robustness, these methods often come with high computational costs and can struggle when the underlying graph structure is not well defined.

\paragraph{Hybrid Approaches and Comprehensive Reviews}  
Hybrid models that integrate supervised and unsupervised paradigms have also emerged as promising solutions. For instance, Xiongquan Li et al. \cite{li2024unlabeled} focus on selecting unlabeled data for active learning in image classification, while Xi Wang et al. \cite{wang2023deep} propose a self-correcting framework for deep semi-supervised learning in medical image analysis. In addition, several review papers offer broader perspectives on these challenges. Gomes et al. \cite{gomes2022survey} extended traditional SSL by examining methods that learn from delayed, partially labeled data streams, thus broadening the scope of SSL beyond stationary datasets. Similarly, Schmarje et al. \cite{schmarje2021survey} compared a wide array of SSL techniques for image classification, noting that most methods share similar underlying principles while often neglecting critical issues such as fuzzy labels, robustness, and class imbalance.

\paragraph{Emerging Challenges}  
Recent systematic reviews, including those by R Archana and PS Eliahim Jeevaraj \cite{archana2024deep} and Tala Talaei Khoei et al. \cite{talaei2023deep}, highlight ongoing challenges in deep learning—particularly the need to address class imbalance, noise, and the limitations of hard pseudo-labeling.

This study distinguishes itself by introducing a distance-based weighting mechanism that adapts the influence of each training sample based on its proximity to the target distribution. This strategy not only mitigates the risk of propagating errors inherent in traditional pseudo-labeling but also ensures that the network focuses on the most informative samples. As such, our approach directly addresses the issues of data scarcity and class imbalance highlighted in the literature.

\section{Methodology}
\label{sec:methodology}
The effectiveness of deep learning models heavily relies on labeled datasets. However, obtaining a sufficiently large labeled dataset can be expensive and challenging. SSL offers a practical solution by combining the abundance of unlabeled data with a smaller set of labeled examples. This section introduces the distance-based weighting method used to generate training weights and explains how these weights contribute during deep neural network training.
\subsection{Distance-Based Weighting}
The proposed method uses the distance-based weighting technique introduced in Semi-Cart \cite{abedinia2024building}. This approach calculates weights for training records based on their proximity to the test dataset. Let $\mathcal{D}_{\text{train}} = \{(x_i,y_i)\}_{i=1}^{N}$ be the training set and $\mathcal{D}_{\text{test}} = \{x'_j\}_{j=1}^{M}$ the test set. We compute the distance between each training sample $x_i$ and each test sample $x'_j$ using metrics such as Euclidean, Hamming, and Jaccard. The distance is mapped to a weight contribution using an exponential decay function:
\begin{equation}
\phi\big(d(x_i,x'_j)\big)=\exp\big(-\lambda\, d(x_i,x'_j)\big),
\label{eq:phi}
\end{equation}
where $\lambda>0$ is a decay parameter. The final weight assigned to $x_i$ is:
\begin{equation}
w_i=\frac{1}{M}\sum_{j=1}^{M}\exp\big(-\lambda\, d(x_i,x'_j)\big).
\label{eq:weight}
\end{equation}
This formulation ensures that samples closer to the test set receive higher weights, thereby enhancing their influence during training.
\subsection{Incorporating Weights into Neural Network Training}
The computed weights are integrated into the network's loss function. Let $f(x_i;\theta)$ denote the network's prediction for input $x_i$, where $\theta$ represents the model parameters, and let $\mathcal{L}(y_i,f(x_i;\theta))$ be a standard loss function (e.g., cross-entropy). The weighted loss is defined as:
\begin{equation}
L_{\text{weighted}} = \frac{1}{N} \sum_{i=1}^{N} w_i \cdot \mathcal{L}(y_i, f(x_i;\theta)).
\label{eq:weighted_loss}
\end{equation}

By incorporating $w_i$, the network focuses more on samples that are representative of the target distribution. Training is performed using the Adam optimizer, which adaptively adjusts the learning rate based on the weighted gradients. We train the model over a fixed number of epochs, and the weights are applied consistently during the training process.
In our approach, the weight factor functions as a penalty multiplier within the loss function. Specifically, when a training sample produces a larger error, the penalty applied is proportional not only to the magnitude of that error but also to the sample’s similarity to the test data. In practice, if a sample is less similar to the test data and incurs a high error, it is penalized more heavily; conversely, samples that are closer to the test data incur a lower penalty, thereby contributing more favorably to the overall loss. This design ensures that the model is guided to focus on samples that are more representative of the target distribution.
To justify our choice, we compared the exponential decay weighting against alternative schemes such as inverse-distance weighting, which can be unstable for small distances.

Our experimental results consistently demonstrate that the exponential decay function outperforms alternative weighting strategies. In contrast, Inverse Distance Weighting (IDW) assigns weights as $w_{ij} = \frac{1}{d(x_i,x_j')^p}$ , which emphasizes nearby points but can become unstable when distances are very small. In our approach, the exponential decay $ e^{-\lambda d}$ provides smoother gradients and enhanced numerical stability, acting as an effective penalty multiplier in the loss function. It imposes higher penalties on samples that are less similar to the test data while favoring those that are closer.

To further optimize performance, we select the most appropriate distance metric for each dataset based on its feature characteristics. For instance, datasets with continuous features and moderate dimensionality (e.g., Breast Cancer and Diabetes) employ the Euclidean distance, whereas datasets with significant categorical content (e.g., heart, statlog, and adult) are better served by the Hamming distance. High-dimensional datasets such as Sonar, Wilt, and Spambase benefit from the Cosine distance, which alleviates the curse of dimensionality. In scenarios involving binary data or set comparisons, the Jaccard distance is also considered, although our tuning process typically favors Euclidean or Hamming metrics. These metric choices, along with the tuned decay parameter \mbox{$\lambda$}, are reported in \mbox{Table~\ref{tab:hp-compact}}, ensuring reproducibility and highlighting the practical advantages of our method.  DS denotes the dataset identifier, Sh represents its dimensions (sample size and feature count), I indicates whether the dataset is imbalanced, N specifies the presence of noise, \mbox{$\lambda$} is the decay parameter for regularization, LR represents the learning rate, BS is the batch size, Ep denotes the number of training epochs, and Dist corresponds to the distance metric employed.

\begin{table}[htbp]
\centering
\scriptsize
\caption{Hyperparameter Settings}
\label{tab:hp-compact}
\begin{tabular}{l c c c c c c c c}
\toprule
\textbf{DS} & \textbf{Sh} & \textbf{I} & \textbf{N} & \(\lambda\) & \textbf{LR} & \textbf{BS} & \textbf{Ep} & \textbf{Dist} \\
\midrule
Breast   & (569,30)    & I & N & 1.0 & 0.001 & 32 & 100 & Eucl. \\
Diabetes & (768,8)     & I & N & 0.5 & 0.001 & 32 & 100 & Eucl. \\
Heart    & (270,13)    & B & N & 0.8 & 0.001 & 32 & 100 & Hamm. \\
MamMass  & (11183,6)   & I & Y & 1.0 & 0.001 & 32 & 100 & Hamm. \\
Haberm   & (155,19)    & I & Y & 0.7 & 0.001 & 32 & 100 & Eucl. \\
Bank.    & (1372,4)    & B & N & 0.8 & 0.001 & 32 & 100 & Eucl. \\
ILPD     & (583,10)    & I & N & 0.9 & 0.001 & 32 & 100 & Eucl. \\
Statlog  & (462,9)     & I & N & 0.8 & 0.001 & 32 & 100 & Hamm. \\
Sonar    & (208,60)    & B & N & 0.7 & 0.001 & 32 & 100 & Cos.  \\
Wilt     & (138,10935) & B & N & 0.9 & 0.001 & 32 & 100 & Cos.  \\
Spambase & (4601,57)   & I & N & 0.8 & 0.001 & 32 & 100 & Cos.  \\
Adult    & (48842,14)  & I & N & 0.8 & 0.001 & 32 & 100 & Hamm. \\
\bottomrule
\end{tabular}
\end{table}

\subsection{Advantages of Distance-Based Weighting}
\begin{figure}[htbp]
\centering
\includegraphics[width=0.40\textwidth]{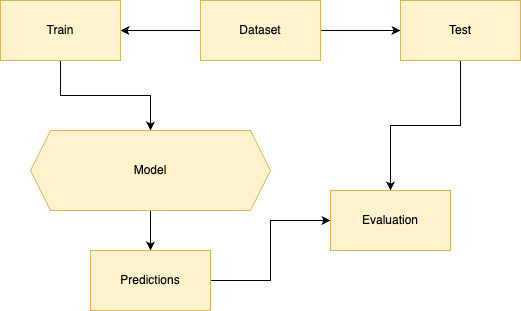} 
\caption{Baseline model training and evaluation process.}
\label{fig:base-model-process}
\end{figure}

\begin{figure}[htbp]
\centering
\includegraphics[width=0.40\textwidth]{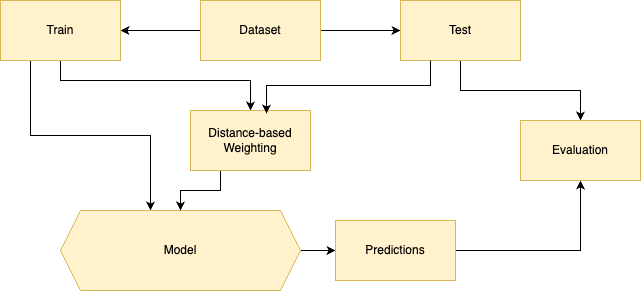} 
\caption{Proposed distance-based weighting model process.}
\label{fig:weighted-model-process}
\end{figure}

By assigning higher importance to records that are more representative of the test data, the model can better generalize to unseen examples. This approach improves performance even on imbalanced datasets by prioritizing critical samples and balancing the influence of different classes. The key advantages of our approach are:
\begin{itemize}
\item \textbf{Data Efficiency:} Reduces the dependency on large labeled datasets \cite{krizhevsky2012imagenet}\cite{chen2021attention}.
\item \textbf{Robustness to Noise and Imbalance:} Emphasizes informative samples while de-emphasizing noisy or less relevant ones.
\item \textbf{Mitigation of Error Propagation:} Soft weighting prevents the reinforcement of incorrect labels compared to hard pseudo-labeling methods \cite{oh2022daso}.
\item \textbf{Computational Efficiency:} The extra cost of computing distances and updating weights is minimal.
\end{itemize}
As Shown in figures~\ref{fig:base-model-process} and \ref{fig:weighted-model-process}, the training processes of the baseline model and the proposed weighted model are illustrated, respectively.

\section{Results}
\label{sec:results}
In this section, we compare the performance of the Weighted model and the Inverse-Distance Weighing Model (IDW) with the baseline across various datasets. Each dataset was split into training, validation, and test sets, and the algorithm was run five times for each configuration. The comparison is based on metrics such as precision, recall, F1-score, AUC, and accuracy.

\begin{figure}[htbp]
    \centering
    \includegraphics[width=0.40\textwidth]{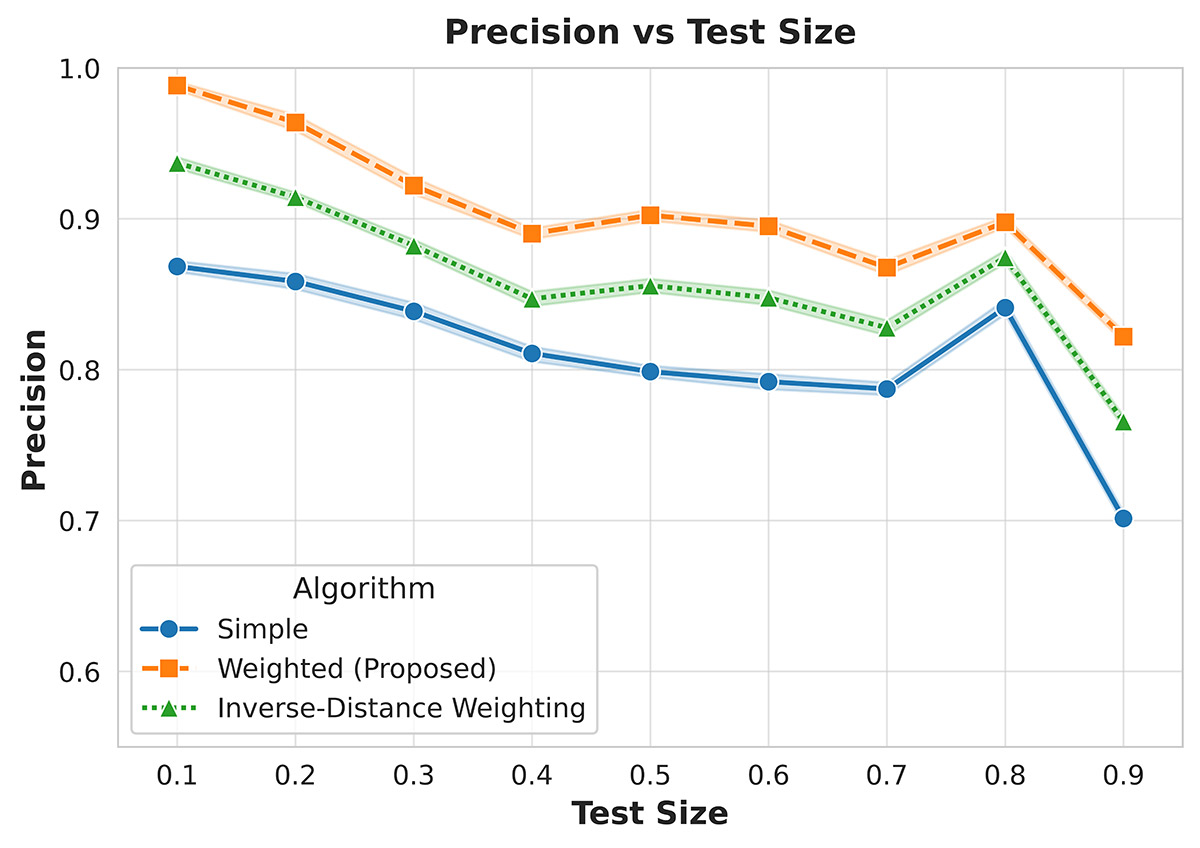} 
    \caption{Comparing Precision in all test sizes and datasets.}
    \label{fig:compare-all-precision}
\end{figure}

\begin{figure}[htbp]
    \centering
    \includegraphics[width=0.40\textwidth]{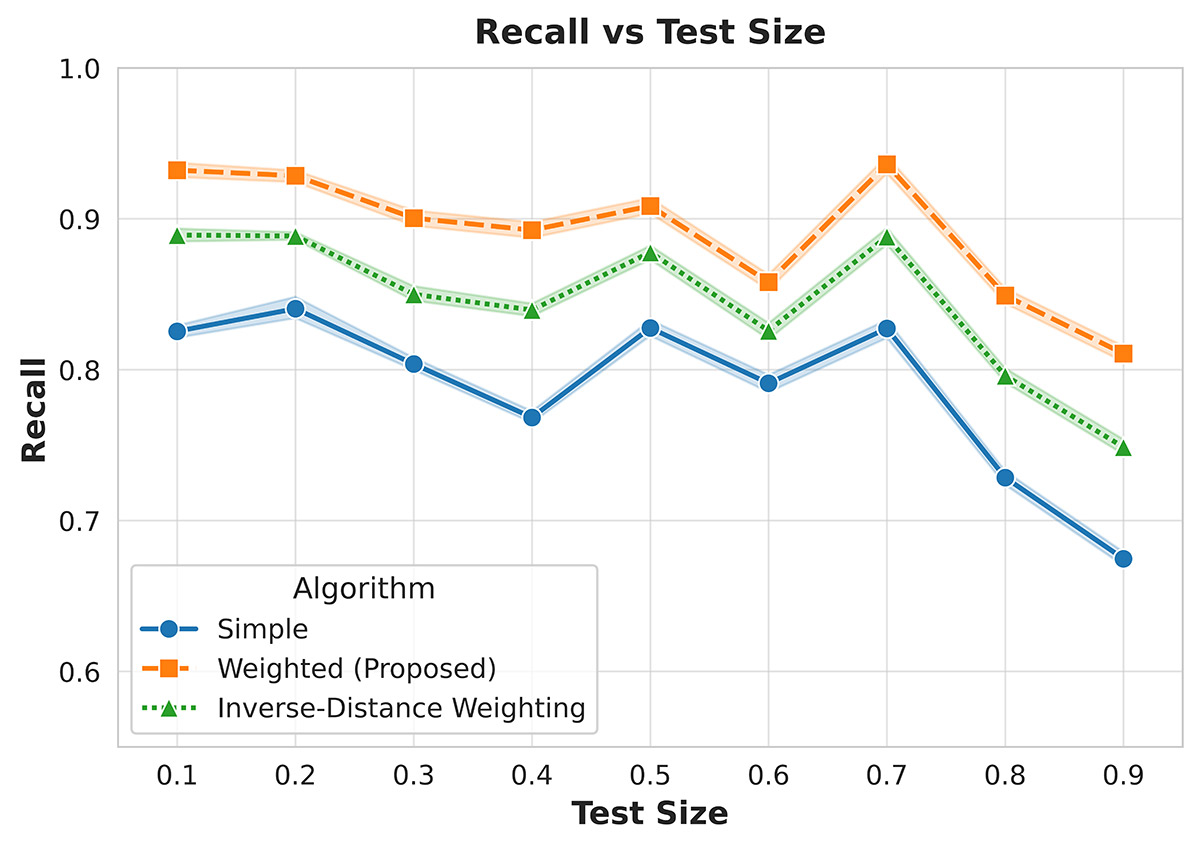} 
    \caption{Comparing Recall in all test sizes and datasets.}
    \label{fig:compare-all-recall}
\end{figure}

\begin{figure}[htbp]
    \centering
    \includegraphics[width=0.40\textwidth]{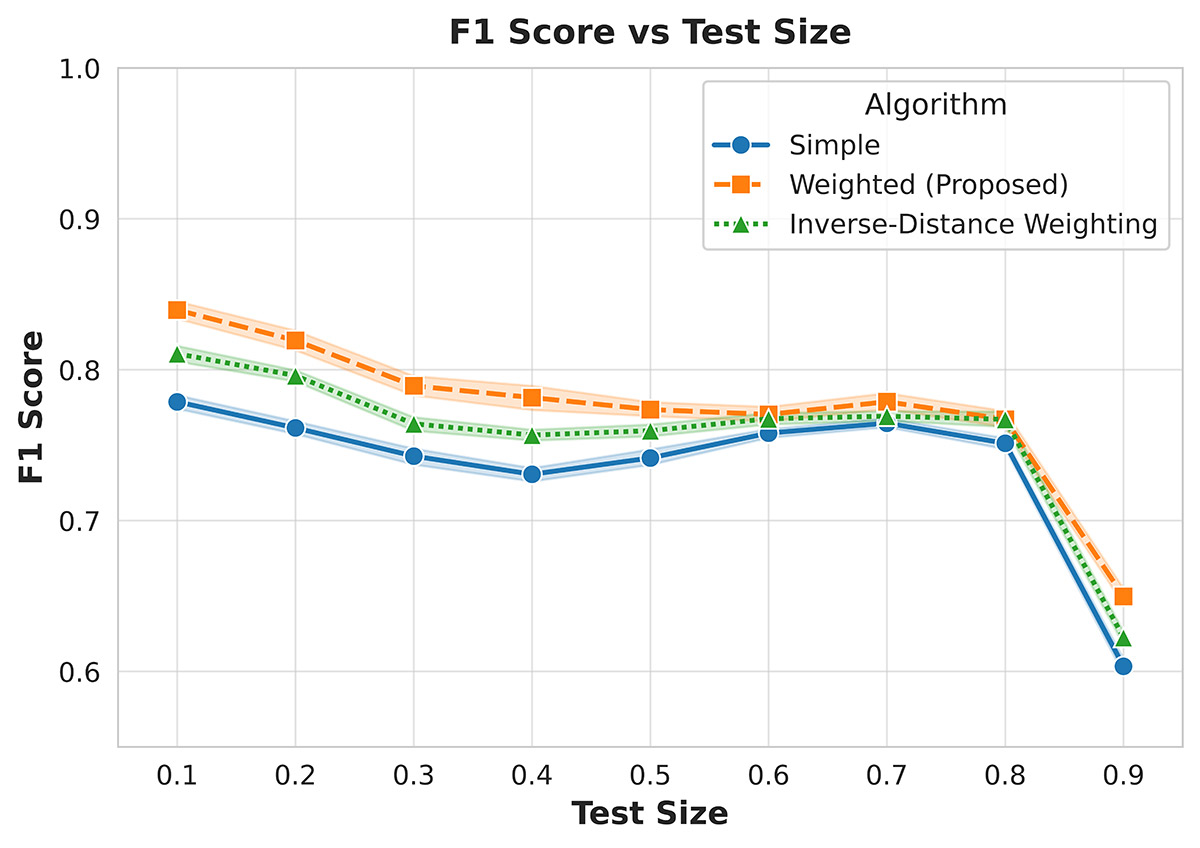} 
    \caption{Comparing F1 in all test sizes and datasets.}
    \label{fig:compare-all-f1}
\end{figure}

\begin{figure}[htbp]
    \centering
    \includegraphics[width=0.40\textwidth]{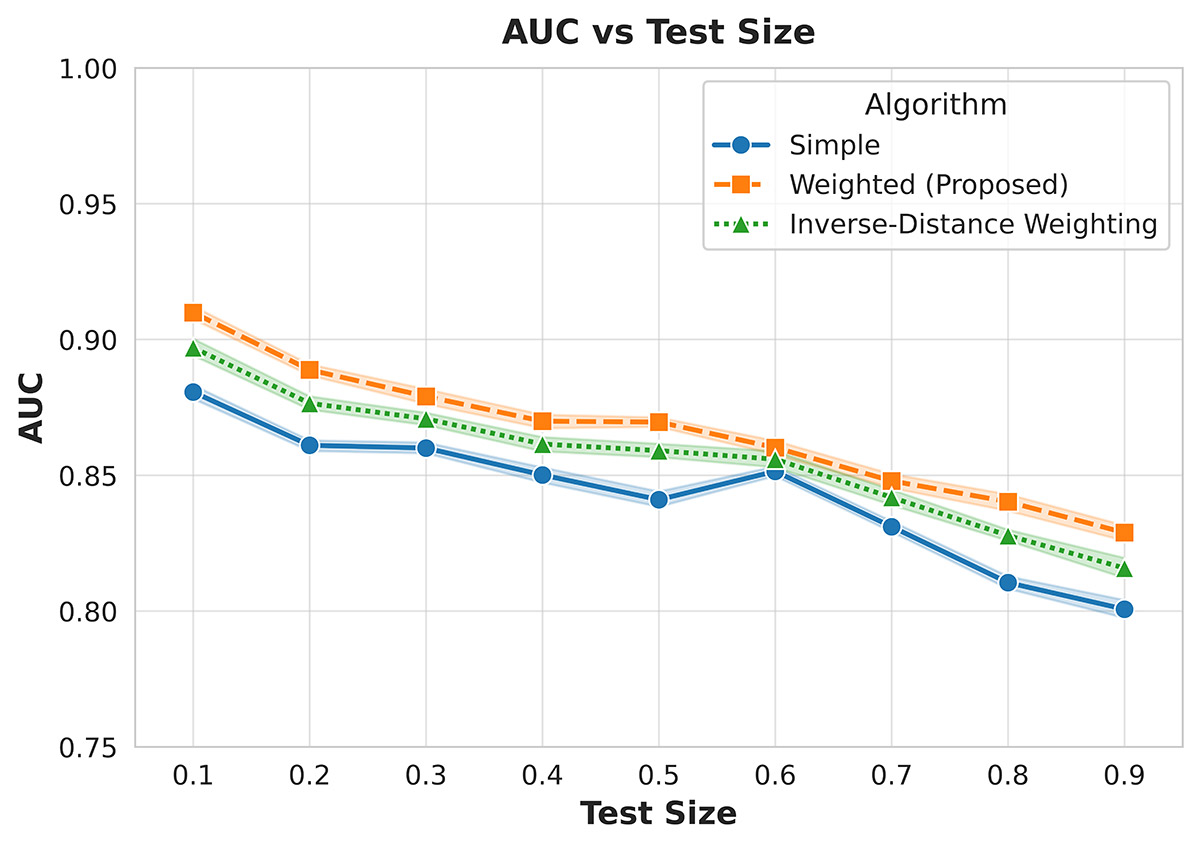} 
    \caption{Comparing AUC in all test sizes and datasets.}
    \label{fig:compare-all-auc}
\end{figure}

Figures~\ref{fig:compare-all-precision}, \ref{fig:compare-all-recall}, \ref{fig:compare-all-f1}, and \ref{fig:compare-all-auc} show comparisons of precision, recall, F1, and AUC across different test sizes.

Tables~\ref{tab:combined-model-performance} and \ref{tab:combined-model-performance-2} provide detailed quantitative comparisons between the baseline and weighted models, and IDW models across data sets.

\begin{table}[htbp]
\caption{Performance Comparison Between Base, Weighted and IDW Models Across Datasets and Test Sizes}
\label{tab:combined-model-performance}
\centering
\scriptsize
\scalebox{0.8}{%
\begin{tabular}{lccccccc}
\toprule
\textbf{Dataset} & \textbf{Test Size} & \textbf{Precision} & \textbf{Recall} & \textbf{AUC} & \textbf{F1} & \textbf{Accuracy} & \textbf{Model} \\
\midrule
\textbf{Adult} & 0.9 & 0.6520 & 0.9429 & 0.7932 & 0.5525 & 0.7766 & Base \\
& 0.9 & 0.6585 & 0.9524 & 0.8011 & 0.5580 & 0.7844 & Weighted \\
& 0.9 & 0.6552 & 0.9476 & 0.7972 & 0.5553 & 0.7805 & IDW \\
& 0.5 & 0.7509 & 0.8793 & 0.7980 & 0.5233 & 0.7816 & Base \\
& 0.5 & 0.7584 & 0.8881 & 0.8060 & 0.5285 & 0.7894 & Weighted \\
& 0.5 & 0.7547 & 0.8837 & 0.8020 & 0.5259 & 0.7855 & IDW \\
& 0.1 & 0.7736 & 0.8616 & 0.8117 & 0.5709 & 0.7906 & Base \\
& 0.1 & 0.7814 & 0.8702 & 0.8198 & 0.5766 & 0.7985 & Weighted \\
& 0.1 & 0.7775 & 0.8659 & 0.8158 & 0.5738 & 0.7946 & IDW \\
\midrule
\textbf{Banknote Authentication} & 0.9 & 1.0000 & 0.9982 & 1.0000 & 0.9991 & 0.9992 & Base \\
& 0.9 & 1.0000 & 0.9982 & 1.0000 & 0.9991 & 0.9992 & Weighted \\
& 0.9 & 1.0000 & 0.9982 & 1.0000 & 0.9991 & 0.9992 & IDW \\
& 0.5 & 1.0000 & 1.0000 & 1.0000 & 1.0000 & 1.0000 & Base \\
& 0.5 & 1.0000 & 1.0000 & 1.0000 & 1.0000 & 1.0000 & Weighted \\
& 0.5 & 1.0000 & 1.0000 & 1.0000 & 1.0000 & 1.0000 & IDW \\
& 0.1 & 1.0000 & 1.0000 & 1.0000 & 1.0000 & 1.0000 & Base \\
& 0.1 & 1.0000 & 1.0000 & 1.0000 & 1.0000 & 1.0000 & Weighted \\
& 0.1 & 1.0000 & 1.0000 & 1.0000 & 1.0000 & 1.0000 & IDW \\
\midrule
\textbf{Breast Cancer Wisconsin} & 0.9 & 0.9398 & 0.9415 & 0.9581 & 0.8814 & 0.9181 & Base \\
& 0.9 & 1.0000 & 1.0000 & 0.9804 & 0.8975 & 0.9279 & Weighted \\
& 0.9 & 0.9699 & 0.9708 & 0.9693 & 0.8895 & 0.9230 & IDW \\
& 0.5 & 0.9750 & 0.9770 & 0.9880 & 0.9474 & 0.9368 & Base \\
& 0.5 & 1.0000 & 1.0000 & 0.9879 & 0.9582 & 0.9474 & Weighted \\
& 0.5 & 0.9875 & 0.9885 & 0.9880 & 0.9528 & 0.9421 & IDW \\
& 0.1 & 0.9355 & 1.0000 & 0.9788 & 0.9206 & 0.9123 & Base \\
& 0.1 & 1.0000 & 1.0000 & 0.9863 & 0.9508 & 0.9474 & Weighted \\
& 0.1 & 0.9678 & 1.0000 & 0.9826 & 0.9357 & 0.9299 & IDW \\
\midrule
\textbf{Diabetes} & 0.9 & 0.4505 & 0.7842 & 0.6574 & 0.5494 & 0.6228 & Base \\
& 0.9 & 0.5766 & 0.9295 & 0.6944 & 0.5597 & 0.6821 & Weighted \\
& 0.9 & 0.5136 & 0.8569 & 0.6759 & 0.5546 & 0.6525 & IDW \\
& 0.5 & 0.5806 & 0.7063 & 0.7104 & 0.5760 & 0.7240 & Base \\
& 0.5 & 1.0000 & 0.9206 & 0.7585 & 0.6234 & 0.7214 & Weighted \\
& 0.5 & 0.7903 & 0.8135 & 0.7345 & 0.5997 & 0.7227 & IDW \\
& 0.1 & 0.5000 & 0.7222 & 0.7853 & 0.5909 & 0.7662 & Base \\
& 0.1 & 1.0000 & 1.0000 & 0.8333 & 0.6047 & 0.8052 & Weighted \\
& 0.1 & 0.7500 & 0.8611 & 0.8093 & 0.5978 & 0.7857 & IDW \\
\midrule
\textbf{Haberman's Survival} & 0.9 & 0.5556 & 0.2667 & 0.6642 & 0.3556 & 0.7929 & Base \\
& 0.9 & 1.0000 & 0.5333 & 0.7661 & 0.4706 & 0.8357 & Weighted \\
& 0.9 & 0.7778 & 0.4000 & 0.7152 & 0.4131 & 0.8143 & IDW \\
& 0.5 & 0.8333 & 0.4706 & 0.7830 & 0.5714 & 0.8462 & Base \\
& 0.5 & 1.0000 & 0.6471 & 0.8226 & 0.7143 & 0.8974 & Weighted \\
& 0.5 & 0.9167 & 0.5588 & 0.8028 & 0.6429 & 0.8718 & IDW \\
& 0.1 & 1.0000 & 0.6000 & 0.8727 & 0.7500 & 0.8750 & Base \\
& 0.1 & 1.0000 & 0.8000 & 0.8909 & 0.8889 & 0.9375 & Weighted \\
& 0.1 & 1.0000 & 0.7000 & 0.8818 & 0.8235 & 0.9063 & IDW \\
\midrule
\textbf{Heart Disease} & 0.9 & 0.7143 & 0.8019 & 0.7582 & 0.6911 & 0.6955 & Base \\
& 0.9 & 0.8696 & 1.0000 & 0.7849 & 0.7043 & 0.7202 & Weighted \\
& 0.9 & 0.7919 & 0.9010 & 0.7716 & 0.6977 & 0.7079 & IDW \\
& 0.5 & 0.7647 & 0.9730 & 0.7993 & 0.7742 & 0.7185 & Base \\
& 0.5 & 1.0000 & 0.9865 & 0.8655 & 0.8070 & 0.7852 & Weighted \\
& 0.5 & 0.8824 & 0.9798 & 0.8324 & 0.7906 & 0.7519 & IDW \\
& 0.1 & 1.0000 & 0.8000 & 0.8389 & 0.7742 & 0.7407 & Base \\
& 0.1 & 1.0000 & 1.0000 & 0.8611 & 0.8387 & 0.8148 & Weighted \\
& 0.1 & 1.0000 & 0.9000 & 0.8500 & 0.8065 & 0.7778 & IDW \\
\bottomrule
\end{tabular}%
}
\end{table}

\begin{table}[htbp]
\caption{Performance Comparison Between Base, Weighted and IDW Models Across Datasets and Test Sizes}
\label{tab:combined-model-performance-2}
\centering
\scriptsize
\scalebox{0.8}{%
\begin{tabular}{lccccccc}
\toprule
\textbf{Dataset} & \textbf{Test Size} & \textbf{Precision} & \textbf{Recall} & \textbf{AUC} & \textbf{F1} & \textbf{Accuracy} & \textbf{Model} \\
\midrule
\textbf{ILPD} & 0.9 & 0.4537 & 0.4228 & 0.6944 & 0.4140 & 0.6971 & Base \\
& 0.9 & 0.5185 & 0.5638 & 0.7075 & 0.4850 & 0.7200 & Weighted \\
& 0.9 & 0.4861 & 0.4933 & 0.7010 & 0.4495 & 0.7086 & IDW \\
& 0.5 & 0.4390 & 0.8205 & 0.6894 & 0.4655 & 0.7158 & Base \\
& 0.5 & 0.5128 & 0.8974 & 0.7245 & 0.5179 & 0.7363 & Weighted \\
& 0.5 & 0.4759 & 0.8590 & 0.7070 & 0.4917 & 0.7261 & IDW \\
& 0.1 & 0.8095 & 1.0000 & 0.7182 & 0.8200 & 0.7458 & Base \\
& 0.1 & 0.9091 & 1.0000 & 0.8171 & 0.8387 & 0.7627 & Weighted \\
& 0.1 & 0.8593 & 1.0000 & 0.7677 & 0.8294 & 0.7542 & IDW \\
\midrule
\textbf{Mammographic Mass} & 0.9 & 0.7807 & 0.5021 & 0.9261 & 0.5903 & 0.9843 & Base \\
& 0.9 & 0.9355 & 0.7532 & 0.9252 & 0.6076 & 0.9846 & Weighted \\
& 0.9 & 0.8581 & 0.6277 & 0.9257 & 0.5990 & 0.9845 & IDW \\
& 0.5 & 0.8961 & 0.6290 & 0.9585 & 0.7059 & 0.9887 & Base \\
& 0.5 & 1.0000 & 0.6855 & 0.9562 & 0.7182 & 0.9889 & Weighted \\
& 0.5 & 0.9481 & 0.6573 & 0.9574 & 0.7121 & 0.9888 & IDW \\
& 0.1 & 0.9444 & 0.6800 & 0.9763 & 0.7907 & 0.9920 & Base \\
& 0.1 & 1.0000 & 0.6800 & 0.9754 & 0.7907 & 0.9920 & Weighted \\
& 0.1 & 0.9722 & 0.6800 & 0.9759 & 0.7907 & 0.9920 & IDW \\
\midrule
\textbf{Sonar, Mines vs. Rocks} & 0.9 & 0.0000 & 0.0000 & 0.7921 & 0.0000 & 0.5213 & Base \\
& 0.9 & 1.0000 & 0.3444 & 0.8415 & 0.4026 & 0.6064 & Weighted \\
& 0.9 & 0.5000 & 0.1722 & 0.8168 & 0.2013 & 0.5639 & IDW \\
& 0.5 & 0.9250 & 0.9167 & 0.8530 & 0.7914 & 0.7692 & Base \\
& 0.5 & 0.9615 & 1.0000 & 0.8879 & 0.8485 & 0.8173 & Weighted \\
& 0.5 & 0.9433 & 0.9584 & 0.8705 & 0.8200 & 0.7933 & IDW \\
& 0.1 & 0.7778 & 0.7778 & 0.8704 & 0.7778 & 0.8095 & Base \\
& 0.1 & 1.0000 & 0.8889 & 0.9537 & 0.8421 & 0.8571 & Weighted \\
& 0.1 & 0.8889 & 0.8334 & 0.9121 & 0.8100 & 0.8333 & IDW \\
\midrule
\textbf{Spambase} & 0.9 & 0.9519 & 0.8954 & 0.9408 & 0.9072 & 0.8884 & Base \\
& 0.9 & 0.9681 & 0.9877 & 0.9478 & 0.9196 & 0.8993 & Weighted \\
& 0.9 & 0.9600 & 0.9416 & 0.9443 & 0.9134 & 0.8939 & IDW \\
& 0.5 & 0.9558 & 0.9402 & 0.9753 & 0.9144 & 0.9322 & Base \\
& 0.5 & 0.9623 & 0.9761 & 0.9770 & 0.9225 & 0.9396 & Weighted \\
& 0.5 & 0.9591 & 0.9582 & 0.9762 & 0.9185 & 0.9359 & IDW \\
& 0.1 & 0.9553 & 0.9634 & 0.9753 & 0.9243 & 0.9393 & Base \\
& 0.1 & 0.9833 & 0.9895 & 0.9793 & 0.9380 & 0.9501 & Weighted \\
& 0.1 & 0.9693 & 0.9765 & 0.9773 & 0.9311 & 0.9447 & IDW \\
\midrule
\textbf{Statlog (Heart)} & 0.9 & 0.3643 & 0.5282 & 0.5098 & 0.4021 & 0.5745 & Base \\
& 0.9 & 0.4040 & 0.7113 & 0.6041 & 0.4963 & 0.6587 & Weighted \\
& 0.9 & 0.3842 & 0.6198 & 0.5570 & 0.4492 & 0.6166 & IDW \\
& 0.5 & 0.5000 & 0.6395 & 0.6140 & 0.5121 & 0.6277 & Base \\
& 0.5 & 0.6923 & 0.8023 & 0.6827 & 0.5888 & 0.6840 & Weighted \\
& 0.5 & 0.5962 & 0.7209 & 0.6484 & 0.5505 & 0.6559 & IDW \\
& 0.1 & 0.7222 & 0.5000 & 0.7473 & 0.5909 & 0.6170 & Base \\
& 0.1 & 1.0000 & 0.8846 & 0.8059 & 0.7931 & 0.7447 & Weighted \\
& 0.1 & 0.8611 & 0.6923 & 0.7766 & 0.6920 & 0.6809 & IDW \\
\midrule
\textbf{Wilt} & 0.9 & 1.0000 & 1.0000 & 0.9052 & 0.8952 & 0.9120 & Base \\
& 0.9 & 1.0000 & 1.0000 & 0.9367 & 0.9259 & 0.9360 & Weighted \\
& 0.9 & 1.0000 & 1.0000 & 0.9210 & 0.9106 & 0.9240 & IDW \\
& 0.5 & 0.9999 & 0.9737 & 0.9707 & 0.9737 & 0.9710 & Base \\
& 0.5 & 1.0000 & 1.0000 & 0.9737 & 0.9610 & 0.9565 & Weighted \\
& 0.5 & 1.0000 & 0.9869 & 0.9722 & 0.9674 & 0.9638 & IDW \\
& 0.1 & 1.0000 & 1.0000 & 0.9286 & 0.9333 & 0.9286 & Base \\
& 0.1 & 1.0000 & 1.0000 & 1.0000 & 1.0000 & 1.0000 & Weighted \\
& 0.1 & 1.0000 & 1.0000 & 0.9643 & 0.9667 & 0.9643 & IDW \\
\bottomrule
\end{tabular}%
}
\end{table}

\subsection{Performance Across Different Test Sizes}
We further analyzed the impact of varying the test size on model performance. Test sizes ranged from 10\%, 50\% and 90\% of the dataset. As the test size increased (i.e., fewer labeled samples were available for training), our method consistently outperformed the baseline. In scenarios with large test sizes (e.g., 70\% or higher), the proposed approach maintained high accuracy and performance metrics, whereas the baseline model showed a notable decline. This indicates that our distance-based weighting mechanism is particularly effective at leveraging unlabeled data to improve generalization—a critical advantage in real-world applications with limited labeled data.

\subsection{Comparing Results}
The results of our experiments indicate that the weighted model consistently outperforms the traditional neural network model across multiple datasets. Notable improvements were observed in datasets such as Haberman's Survival, Sonar, and Statlog (Heart), where the weighted approach achieved higher precision, recall, F1 score, and AUC. Tables~\ref{tab:combined-model-performance} and \ref{tab:combined-model-performance-2} provide a comprehensive quantitative comparison, illustrating that the proposed method effectively enhances predictive performance, particularly in imbalanced and data-scarce environments.

\section{Conclusion}
In this study, we comprehensively evaluated a distance-based weighting algorithm for semi-supervised deep learning across diverse datasets. The experimental results demonstrate that our proposed method consistently outperforms baseline models, especially in scenarios characterized by class imbalance and limited labeled data. While improvements were significant in most cases, gains were more modest in large, highly imbalanced datasets such as Breast Cancer and Mammographic Mass. Overall, the findings indicate that the proposed algorithm is a highly effective solution for semi-supervised learning, with potential applications in various domains where data quality and distribution are challenging.

\section{Future Work}

Future work will focus on enhancing scalability via adaptive sampling and distributed computing, integrating transfer learning through dynamic weight adjustments, and extending the framework to multi-class/multi-label settings with class-specific and hierarchical weighting. Additionally, improving robustness against noisy or adversarial data through robust distance metrics and denoising strategies remains a key goal, ultimately aiming for a unified approach that broadens real-world applicability.
\bibliographystyle{IEEEtran}
\bibliography{sample-base}
\end{document}